\documentclass{article}

\usepackage{microtype}
\usepackage{graphicx}
\usepackage{subfigure}
\usepackage{tikz}
\usetikzlibrary{arrows,automata}
\usepackage{amsfonts}
\usepackage{booktabs} 

\usepackage{hyperref}

\usepackage[accepted]{icml2020}

\icmltitlerunning{Structure Mapping for Transferability of Causal Models}

\begin{document}

\twocolumn[
\icmltitle{Structure Mapping for Transferability of Causal Models}



\icmlsetsymbol{equal}{*}

\begin{icmlauthorlist}
\icmlauthor{Purva Pruthi}{umass}
\icmlauthor{Javier González}{equal,msft}
\icmlauthor{Xiaoyu Lu}{amzn}
\icmlauthor{Madalina Fiterau}{umass}

\end{icmlauthorlist}

\icmlaffiliation{umass}{College of Information and Computer Sciences, University of Massachusetts Amherst, USA }
\icmlaffiliation{amzn}{Amazon, Cambridge, United Kingdom }
\icmlaffiliation{msft}{Microsoft Research, Cambridge, United Kingdom }

\icmlcorrespondingauthor{Purva Pruthi}{ppruthi@cs.umass.edu}

\icmlkeywords{Machine Learning, ICML}

\vskip 0.3in
]



\printAffiliationsAndNotice{\textsuperscript{*}Work done while working at Amazon, Cambridge, UK.} 

\begin{abstract}
 Human beings learn causal models and constantly use them to transfer knowledge between similar environments. We use this intuition to design a transfer-learning framework using object-oriented representations to learn the causal relationships between objects. A learned causal dynamics model can be used to transfer between variants of an environment with exchangeable perceptual features among objects but with the same underlying causal dynamics. We adapt continuous optimization for structure learning techniques \cite{zheng2018dags} to explicitly learn the cause and effects of the actions in an interactive environment and transfer to the target domain by categorization of the objects based on causal knowledge. We demonstrate the advantages of our approach in a gridworld setting by combining \textit{causal model-based approach} with \textit{model-free approach} in reinforcement learning.
 \end{abstract}

\section{Introduction}
In reinforcement learning, two kind of approaches are common - \textit{model-free} learning and \textit{model-based} learning \cite{sutton1999reinforcement}. In model-free learning, the agent learns to directly estimate the future reward for different states without building an explicit model of the system while in model-based learning the agent explicitly learns the state transition and reward distributions of the system. In model-based learning, the model can be
prediction-based, where the goal is to accurately predict the future state or it can be causal, where the goal is to explicitly model the variables (causes) responsible for the state transition and rewards. Experimental work in cognitive science has shown that humans and animals use a combination of model-based and model-free algorithms.\cite{dolan2013goals}


Building explicit causal models for the systems provide the advantage of explicit understanding of the dynamics of the system and re-using the causal knowledge to transfer to the variations of the system. For example, while learning a new video game, we first learn about different objects within the game, their perceptual features (e.g. color, shape, material etc.), causal characteristics (like what will happen if you hit a wall) and causal relationships between different objects (e.g. switching on the switch causes bulb to light.) Now, if we are presented with a new variant of the game with different looking objects but similar causal characteristics then we tend to transfer knowledge by mapping the objects seen in the previous game with the objects in the new game based on similar causal behavior. On the other hand, if we are presented with a new variant of the game with similar looking objects but with different causal characteristics, then we need to augment the causal model for those objects with new causal characteristics.

In either case, multiple characterization of the objects in terms of their perceptual features as well as causal properties provide several advantages. First, this characterization allows the flexibility to decouple the perceptual representation of an object from their causal properties, allowing combinatorial generalization over systems with different combination of these features. Second, it allows to build abstract causal knowledge in the form of high-level causal concepts. (e.g. Hitting a wall causes no movement even if the wall were red, blue or black in color). [\cite{Kemp2010LearningTL}].

Object-oriented MDP \cite{diuk2008object} provides an efficient way of modeling an environment as it factorizes the state space over different objects and thus, reducing the large state-space. Also, it provides an abstraction by attributing the effect of the objects to the object's characteristics rather than it's specific location in the state space, which is the case in pixel-based representation (assuming the model is trained on images).

In this work, we combine ideas from object-oriented MDPs \cite{diuk2008object}, structure learning \cite{zheng2018dags}, and categorization as causal reasoning \cite{rehder2003categorization}, \cite{Kemp2010LearningTL}. Our main contribution can be summarized as below:
\icmlitem We formulate and evaluate the learning of the state transition model in OO-MDP as a parametric structure learning problem using continuous optimization of score-based structure learning techniques.
\icmlitem We formulate and evaluate the causal dynamics model combined with model-free approach on a toy gridworld problem as a proof-of-concept.
\icmlitem We formulate transfer learning as a structure mapping problem where we learn mapping from objects in the source domain to the target domain based on the causal structure.
\begin{figure}[ht]
\vskip 0.2in
\begin{center}
\centerline{\includegraphics[width=\columnwidth,scale=0.5]{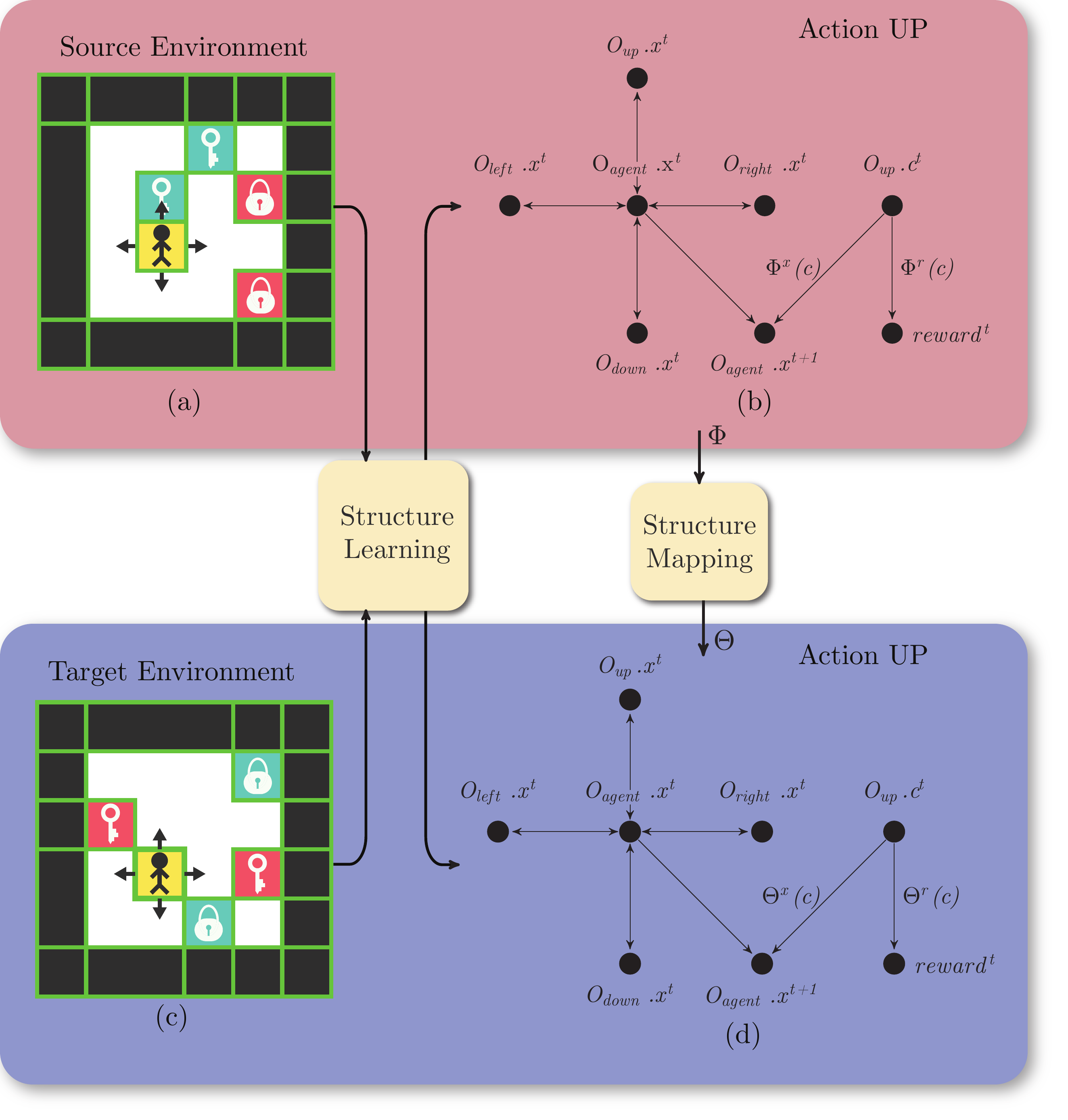}}
\caption{High-level schematic diagram to illustrate the overall approach for transfer learning.
(a) Object based segmentation of source environment consisting of an agent (in yellow), two keys (in blue-green) and two locks (in red). Each object is described by it's attributes \{\textit{x-position} (O.x), \textit{y-position} (O.y), \textit{color} (O.c)\} attributes (b) Causal Dynamics model of source environment for taking action `UP'. (c) Target environment with same rules as source environment but different (inverted) colors of keys and locks. (d) Causal Dynamics model of target environment for taking action `UP'.}
\label{schematic}
\end{center}
\vskip -0.2in
\end{figure}
\section{Preliminaries and Problem Statement}
\subsection{Notation}
The traditional formalism for the reinforcement learning problem is the Markov Decision Process (MDP). An MDP is a five-tuple $(\mathcal{S}, \mathcal{A}, T, \mathcal{R}, \gamma)$, where $\mathcal{S}$ is a set of states, $\mathcal{A}$ is the set of actions, $T(s^{(t+1)}|s^{(t)}, a^{(t)})$ is the probability of transforming from state $s^{(t)} \in \mathcal{S}$ to $s^{(t+1)} \in \mathcal{S}$ after action $a^{(t)} \in \mathcal{A}$, $\mathcal{R}(r^{(t+1)}|s^{(t)}, a^{(t)})$ is the probability of receiving the reward $r^{(t+1)} \in \mathbb{R}$ after executing action $a^{(t)}$ while in state $s^{(t)}$, and $\gamma \in [0,1]$ is the rate at which future rewards are exponentially discounted. In object-oriented MDP, we assume that the state consists of $N$ objects (or entities) and each object has $M$ attributes. $O_i$ refers to the $i^{th}$ object and let $\alpha_{i,j}^{(t)}$ refer to the $j^{th}$ attribute value of the $i^{th}$ object at time $t$. $O^{(t)}_{i}= (\alpha^{(t)}_{i,1}, \dots \alpha^{(t)}_{i,M})$ refers to the state of the $i^{th}$ object at time $t$. The complete state of the MDP at time $t$ is $s^{(t)} = (O^{(t)}_{1} \dots O^{(t)}_N)$. Example of an attribute $\alpha_{ij}$ is \{\textit{color}, \textit{$x$-position}, \textit{$y$-position} etc\}. Attributes which remain constant over time in a given environment characterize the object. (e.g. $color$ characterizes locks, keys, agent and walls in a particular environment.)
\subsection{Problem Statement}
Let's assume that there are two \textit{deterministic} environments, source $S$ and target $T$. In this work, we assume that the two environments have same causal dynamics model but different perceptual features. In $S$, agent can experiment with the environment \textit{more freely} by taking random actions while it is \textit{more expensive} to experiment in $T$. Our goal is to transfer knowledge from $S$ to $T$ with minimum possible interventions in $T$ and learn an optimal policy for $T$.

\subsection{Toy Environment: Triggers}
We provide a proof-of-concept for our approach using a toy-environment (known as Triggers), first introduced in \cite{ferret2019selfattentional}. In this gridworld based environment, the agent can take one of the four possible actions $\{north, south, east \ or \ west\}$. Hitting black walls doesn't change the agent's position and moving in the free space increases or decreases the agent's $x$ or $y$ position by 1, depending on the action. In source environment (Figure \ref{schematic}(a)), green colored boxes represent the keys and red-colored boxes represent the locks. The agent should collect \textbf{all} the keys before it can attempt opening one of the locks. If the agent attempts to open the lock when at least one of the keys is still present, the agent receives a negative reward of $-1$. It receives $+1$ reward for opening each lock successfully. The key boxes disappear when collected and the lock boxes disappear when opened successfully. In the target environment, (Figure \ref{schematic}(d)), we change the colors of the keys and locks. The goal is to transfer the knowledge from source to target by mapping the causal behavior of the objects (keys and doors) with corresponding colors (perceptual features).
\section{Our approach}
Our approach mainly consists of three steps which are also summarized in Figure \ref{schematic}. We assume that the object-oriented representation of the environment is learned from input images using unsupervised visual-object reasoning modules \cite{eslami2016attend} and is already \textit{provided} to us. We learn the causal dynamics model $M^{S}$ of the source environment from the agent's interactions in the source environment using a random policy and use the dynamics model for planning in $S$. The causal dynamics model consists of the causal structure between the object attributes at the current time-step and the next time-step, parametrized by $\Phi$. In the target environment, we start with the assumption that it follows the same dynamics model $M^{S}$. We simulate the initial trajectories for $T_0$ timesteps and also observe the state transition and reward by actually experimenting in the target environment. This will lead to the inaccurate prediction using $M^{S}$ when interacting with some set of objects ${O^T_1 \dots O^T_s}$. We learn a mapping from attributes $\alpha_{ij}'$ of such objects in the target environment $O^T$ to attributes $\alpha_{ij}$ of objects in source environment $O^S$ such that $M^S$ is invariant in $T$.

\subsection{Structure Learning in Object Oriented-MDPs}
The problem of structure learning in OO-MDPs can be formulated as learning a directed acyclic graph $G^{a} = (V,E)$ with object attributes as vertices $V \in \mathbb{R}^{N \times M}$ and $E \in \mathbb{R}^{NM \times NM}$. We assume that the dependencies encoded by DAG represent direct causal dependencies. This assumption is reasonable in our context because we discover the causal graph by performing interventions (random policy) in the environment rather than by using observational data.
Note, that in OO-MDPs each $G^a$ is specific to each value $a$ $\in \mathcal{A}$ of an action because different actions might result in different causes for the state transitions, resulting in different graphs. For example, if an agent takes an `UP' action, then the only attributes corresponding to the object in the `UP' direction might be valid causes for the state transition of the agent. Due to the determinstic and stationary nature of the environment, we obtained one fixed graph $G^a$ for each action $a$.

\cite{zheng2018dags} \cite{zheng2020learning} introduced an approach (NOTEARS) in which structure learning can be formulated as a continuous optimization problem, making the score-based structure learning approaches efficient for high-dimensional input variables.

Let us represent the current state and next state of attributes as $X = \{\alpha_{11}^{(t)},\alpha_{12}^{(t)} \dots \alpha_{NM}^{(t)}, \alpha_{11}^{(t+1)},\alpha_{12}^{(t+1)}, \dots \alpha_{NM}^{(t+1)} \}$ with dimension $d = 2NM$,
Causal generative process of each attribute $X_j$ can be written as: $ X_j = f_{j}(Pa(X_j))$.
We assume that the attributes at time $t+1$ cannot be parents of attributes at time $t$. Though, we allow the edges between the attributes corresponding to same timestamp (e.g. $\alpha_{ip}^{(t)}$ and $\alpha_{jq}^{(t)})$ which are interpreted as relational dependencies rather than a causal dependency (e.g. spatial relationship between $x$-position of the up neighbor and $x$-position of  the agent). Assuming agent as reference, dependence between x-y position of neighboring objects and agent can be encoded as a directed edge.

Our goal is to learn a DAG $G^a(\Phi)$ by learning $f = (f_1, f_2, f_3 \dots f_d)$. We assume that $f_j$ is a function approximated by neural networks using the parameters $\Phi_j$. Thus, the optimization problem becomes
$$ \min_{\Phi} L(\Phi) = \frac{1}{n}\sum_{j=1}^{d} l(x_{j}, f(X, \Phi_j))$$
subject to
$$ h(W(\Phi)) = 0$$
where $W(f) = W(f_1, f_2 \dots f_d) \in \mathbb{R}^{d \times d}$ encodes the graph edges $G^a(\Phi)$, i.e. $[w(f)]_{kj} = \Vert {\frac{\partial f_j}{\partial X_k}}\Vert_{L_2} = 0$ if $X_k \notin Pa(X_j)$ and $h(W(\Phi)) = tr(e^{W \cdot W}) - d$ encodes the acyclicity constraint. Refer \cite{zheng2018dags} and \cite{zheng2020learning} for more details.

Previously, \cite{kansky2017schema} have used a similar approach called Schema Networks for zero-shot learning by building an explicit causal generative model. Our approach to learn causal structure using NOTEARS provides several advantages compared to the integer-programming based structure learning used in Schema Networks. First, the binary representation for attributes used by the schema networks limits the scope of their approach as binary representation makes the approach computationally challenging for high-dimensional attributes. NOTEARS with parametric DAGs, on the other hand allows using continuous and discrete variables for the structure learning. Second, NOTEARS with object-oriented representation can learn state-space abstraction by capturing relational dependencies. For example, if two objects interact when they are adjacent to each other, then the interaction between attributes of different objects can be learned in the functional form of $f_k(j)$ regardless of the absolute positions of the objects.

\subsection{Structure Mapping from Source to Target}
We describe our structure mapping approach in Algorithm \ref{alg:cdqa}. Intuitively, there are two ways to do structure mapping, either by updating $G(\Phi)$ to $G(\Theta)$ or by mapping the source object attributes to target object attributes and categorize objects in the same group if causal behaviors are the same. The first approach requires learning the functional relationships $\Theta$ from scratch, assuming the graph structure remains the same. If we assume underlying causal dynamics remain the same, then we can adopt the second approach where the goal is to learn a mapping from source objects to target objects. (See Algorithm \ref{alg:cdqa})
\begin{algorithm}[tb]
\caption{Structure Mapping from $G(\Phi)$ to $G(\Theta)$}
  \label{alg:cdqa}
\begin{algorithmic}
  \STATE {\bfseries Input:} Causal Dynamics Model $G^{a}(\Phi)$, Optimal $Q$-values $Q^{*}(S,A)$ from source environment
  \REPEAT
  \STATE $S \leftarrow$ Current state in object-oriented representation.
  \STATE $A \leftarrow  \epsilon-greedy(S,Q)$
  \STATE Execute action $A$, observe next state $S'$ and reward $R$
  \STATE $R^{G}, S'^{G} \leftarrow$ $G^{a}(S, \Phi)$
   \IF{$R \neq R^{G}$ OR $S' \neq S'^{G}$}
    \STATE $O^A.attr \leftarrow O^{A}.color$ (Pick attribute which characterize object)
    \STATE Identify $O^S$ such that $\Theta^{reward}(O^A.attr) = \Phi^{reward}(O^S.attr) = R$ and $\Theta^{state}(O^A.attr) = \Phi^{state}(O^S.attr) = S'$
    \STATE Categorize $O^S$ and $O^A$ in same group and map attribute $O^A.attr$ to $O^S.attr$
    \ENDIF
  \UNTIL{$T_0$ steps}
\end{algorithmic}
\end{algorithm}
\section{Experimental Evaluation}
\subsection{Structure Learning}
We trained the structure learning algorithm on the source Trigger environment with different size of the state space (by varying grid's width and height from $5$ to $75$) and different numbers and locations of the keys and locks. Input $X$ is 19-dimensional vector containing (x-position, y-position, color) attributes of agent, four neighbors and reward, and agent's x,y positions at $t+1$. We also include \texttt{num\_keys} as the variable which represents the number of keys present in the environment at each time step.
\begin{figure}[ht]
\vskip -0.1 in
\begin{center}
\centerline{\includegraphics[width=\columnwidth]{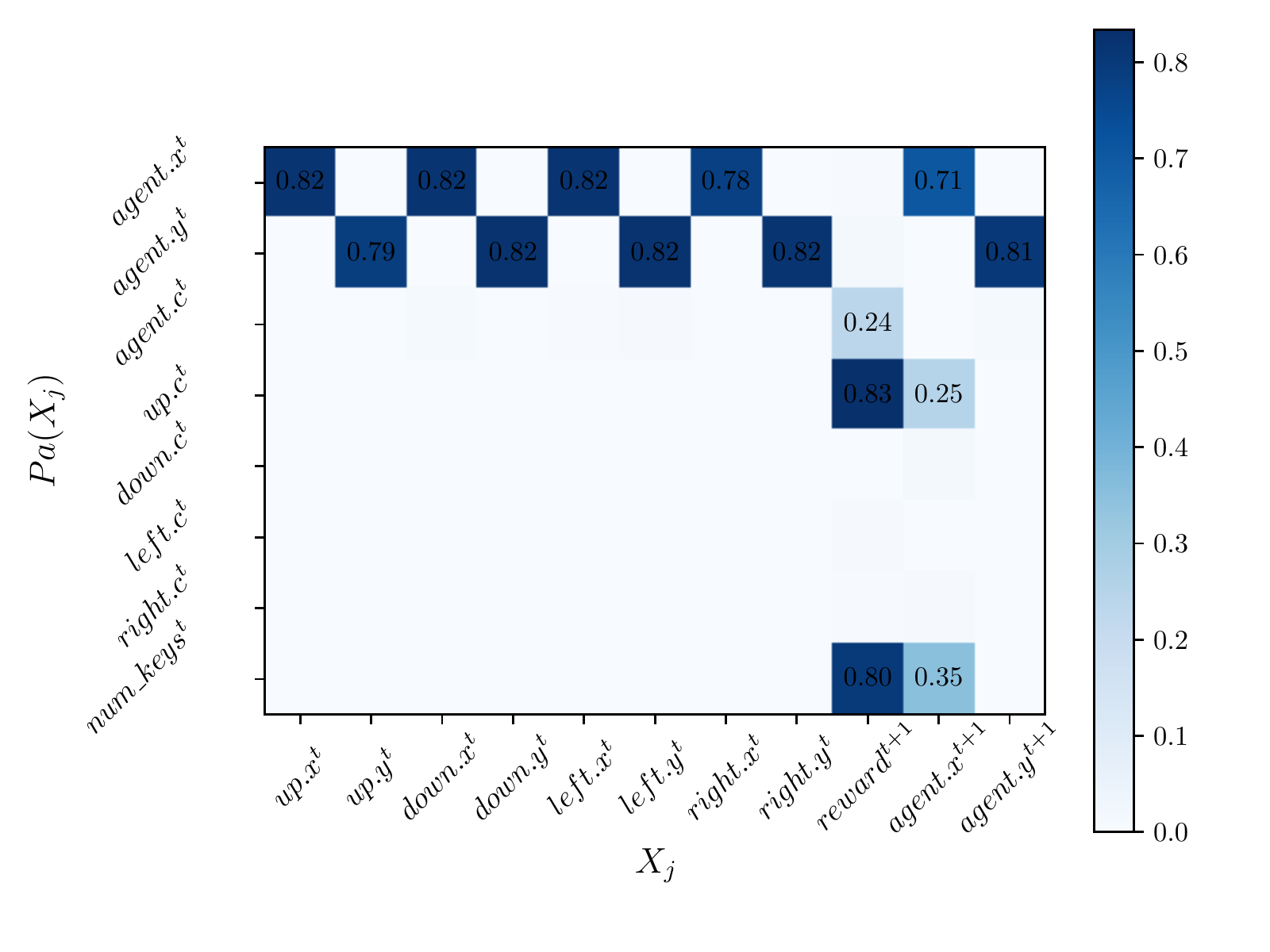}}
\vspace{-4
ex}
\caption{$L_2$ norm $\Vert{\Phi_j(X)}\Vert_{L_2}$ of Weight Values for learned bayesian graph $G^{up}$ for action `UP'. Attributes on y-axis are potential parents (causes) of the corresponding $j_{th}$ attribute on x-axis ($X_j$), with parameters $\Phi_j$. Positive values of weights indicate edge between parents and children.}
\label{structlearning}
\end{center}
\vskip -0.2in
\end{figure}
In Figure \ref{structlearning}, we observe that the attribute future reward $r^{t}$ has total number of keys ($num\_keys$) and color of the `up' neighbor ($up.c^{t}$) as the parents, as expected.  For agent's next positions $x^{t+1}$ and $y^{t+1}$ , it's current x and y positions and color of the `up' neighbor are the causes. We also see edges between x and y positions of the neighbors and the agent due to spatial dependency between them.
\subsection{Agent's Performance in Source Environment}
\begin{figure}[ht]
\vskip -0.15in
\begin{center}
\centerline{\includegraphics[width=\columnwidth]{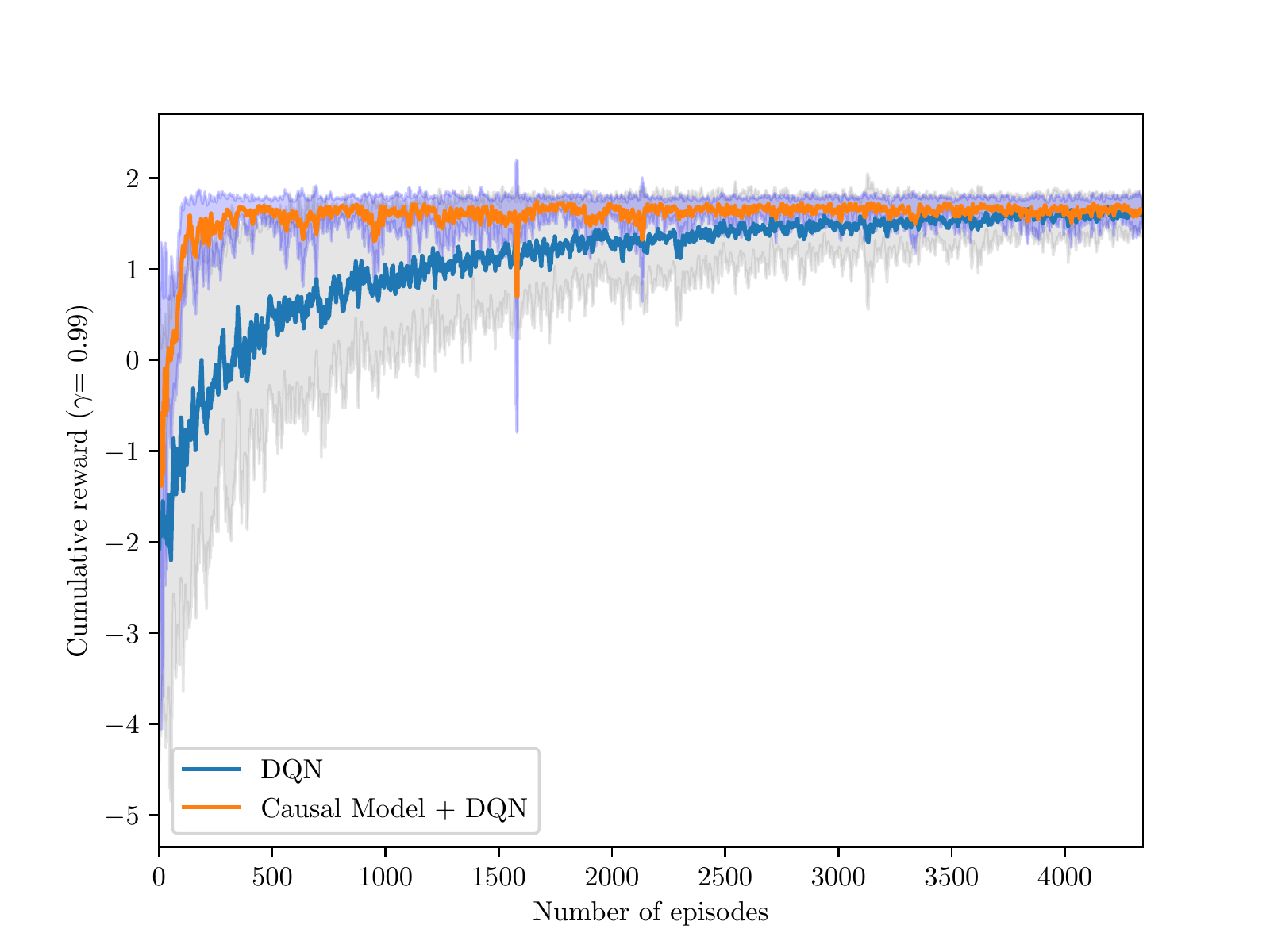}}
\vspace{-2ex}
\caption{Planning using Causal-Dynamics Model}
\label{performance}
\end{center}
\vskip -0.4in
\end{figure}

To evaluate the efficiency of learned causal dynamics model, we use the model to do planning alongwith exploration in the source environment. We use random shooting planning algorithm \cite{nagab2017neural} to select the optimal action sequence. Figure \ref{performance} compares the performance of trained DQN agent and DQN agent alongwith online causal planning agent (Causal Model +DQN). Exploration parameter $\epsilon$ is reduced to 0.05 in DQN after planning steps due to which less than 50\% less random actions were taken by combined model-free and model-based approach. We observe that combined agent converges faster than the purely model-free agent, with less variance. (See Appendix for more details.) Code required to reproduce our experiments is available online at \href{https://github.com/Information-Fusion-Lab-Umass/causal_transfer_learning}
{\texttt{https://github.com/Information-Fusion\\-Lab-Umass/causal\_transfer\_learning}.}
\section{Discussion and Future Work}
In this work, we built an explicit causal dynamics model for deterministic and discrete environments with the goal to transfer the knowledge in domains with similar dynamics. We observe that this is a flexible and interepretable way to model perceptually changing environments. We are in the process of evaluating our structure mapping on Triggers and extending it to real-world environments. The main challenge is in learning object-oriented representation for objects in an unsupervised manner for complex domains with continuous action and state spaces. Limited intervention setting and use of largely available observational data for structure learning are some interesting future directions for our work.


\newpage

\appendix
\section{More Experiment Details}
Code required to reproduce our experiments is available online at \href{https://github.com/Information-Fusion-Lab-Umass/causal_transfer_learning}
{\texttt{https://github.com/Information-\\Fusion-Lab-Umass/causal\_transfer\_learning}.}
\subsection{Structure Learning}
\begin{figure}[htbp!]
\vskip -0.15 in
\begin{center}
\centerline{\includegraphics[width=\columnwidth]{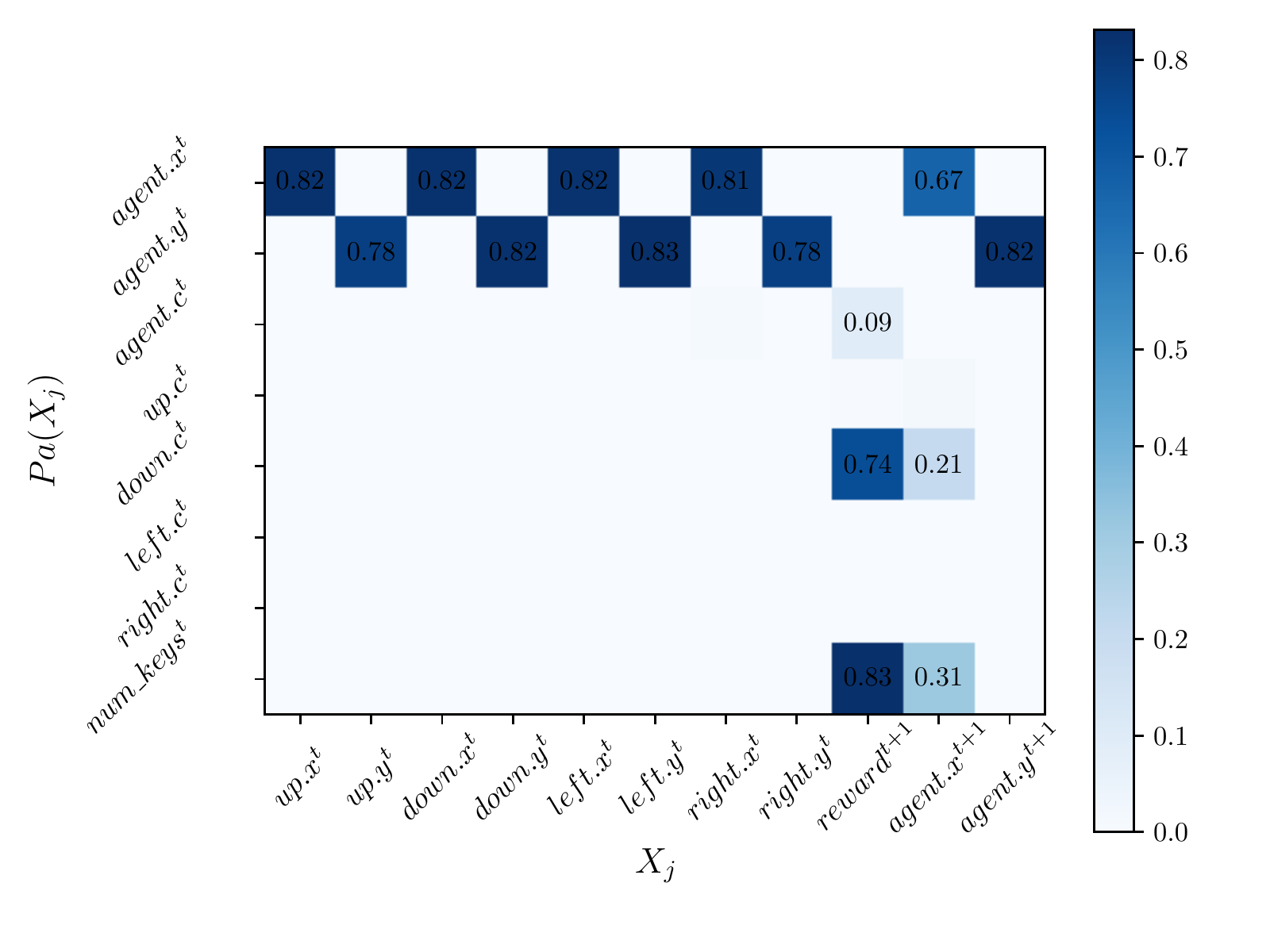}}
\vskip -0.2 in
\caption{$L_2$ norm $\Vert{\Phi_j(X)}\Vert_{L_2}$ of Weight Values for learned bayesian graph $G^{down}$ for action `DOWN'. Attributes on y-axis are potential parents (causes) of the corresponding $j_{th}$ attribute on x-axis ($X_j$), with parameters $\Phi_j$. Positive values of weights indicate edge between parents and children.}
\label{structlearning_down}
\end{center}
\end{figure}

\begin{figure}[ht]
\vskip -0.1 in
\begin{center}
\centerline{\includegraphics[width=\columnwidth]{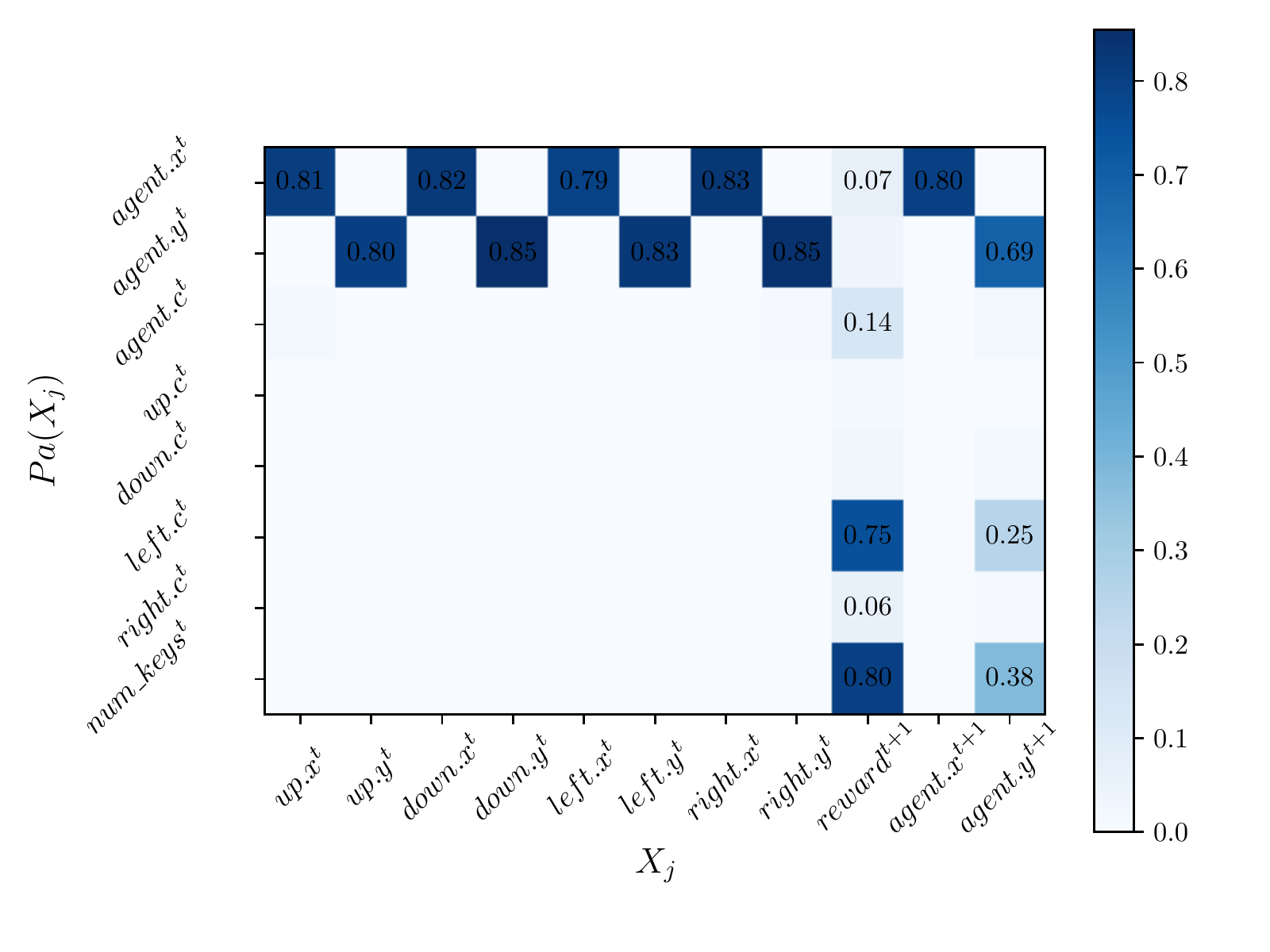}}
\vskip -0.2 in
\caption{$L_2$ norm $\Vert{\Phi_j(X)}\Vert_{L_2}$ of Weight Values for learned bayesian graph $G^{left}$ for action `LEFT'.}
\label{structlearning_left}
\end{center}
\end{figure}

\begin{figure}[htbp!]
\vskip -0.2 in
\begin{center}
\centerline{\includegraphics[width=\columnwidth]{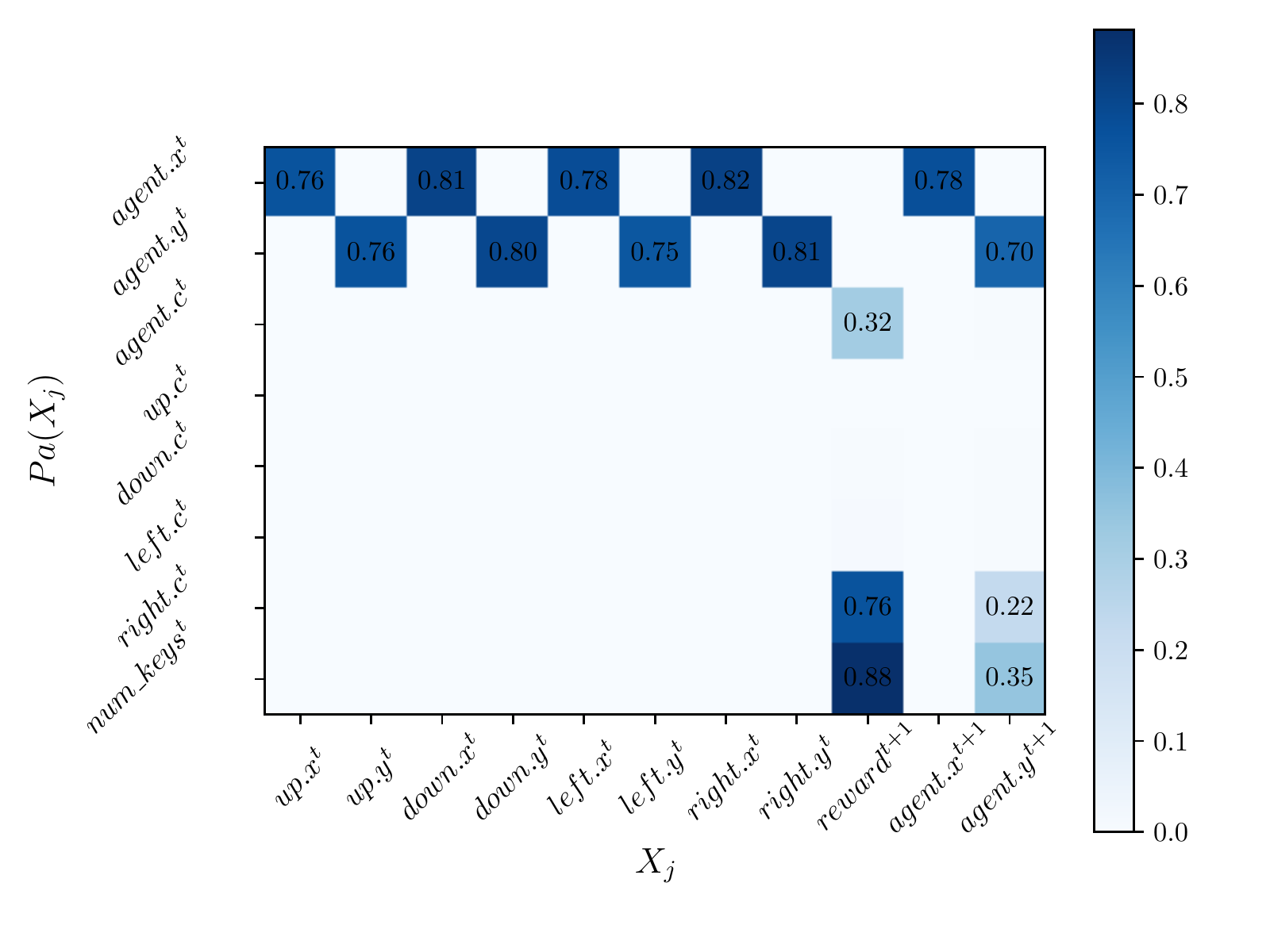}}
\vskip -0.25 in
\caption{$L_2$ norm $\Vert{\Phi_j(X)}\Vert_{L_2}$ of Weight Values for learned bayesian graph $G^{right}$ for action `RIGHT'.}
\label{structlearning_right}
\end{center}
\end{figure}

To learn the causal structure, we trained the NOTEARS non-linear structure learning algorithm on 16000 sample points for each action UP, DOWN, LEFT and RIGHT. For robust learning of the causal model, data is collected using different size of the state space (by varying grid’s width and height from 5 to 75) and different numbers and locations of the keys and locks. We used random policy to generate (state, action, next state, reward) tuple from the environment. Input $X$ is 19-dimensional vector containing (x-position, y-position, color) attributes of agent, four neighbors and reward, and agent's x,y positions at $t+1$. We also include \texttt{num\_keys} as the variable which represents the number of keys present in the environment at each time step. Regularization parameters used for non-linear NOTEARS algorithm were $\lambda_1 = 0.01, \lambda_2 = 0.01 \ and \ \rho = 1.0$. Structure learning results for actions DOWN, LEFT, RIGHT are shown in Figure \ref{structlearning_down}, \ref{structlearning_left}, and \ref{structlearning_right} respectively.
Check \href{https://github.com/xunzheng/notears/tree/master/notears}{\texttt{https://github.com/xunzheng/notears}} for more details on NOTEARS algorithm.

\subsection{Agent's Performance in Source Environment}

\subsubsection{DQN Model Training}
DQN model was trained for total $250000$ steps with initial $3000$ steps as burning period to fill the replay buffer with exploration parameter $\epsilon = 1.0$. From 3000 steps to 250000 steps, $\epsilon$ was decayed linearly from $1.0$ to $0.05$ in the environment at train time. The network obtains same input as structure learning algorithm (19-dimensional object-oriented state vector). We use RMSProp as an optimizer with a base learning rate of 0.00025. We update the target network every 2000 steps.

\subsubsection{Causal Model + DQN Training}
For each action, we use structure learned from NOTEARS algorithm. We use 5000 planning steps with random shooting algorithm to select the 5 random sequences of length 100. Optimal sequence with highest cumulative reward is selected. We only choose the first action from the optimal action sequence and play that action in real environment to train DQN. After 5000 planning steps, we run DQN for 20000 training steps, with fixed exploration parameter $\epsilon = 0.05$.

\end{document}